\documentclass[conference]{IEEEtran}
\IEEEoverridecommandlockouts
\usepackage{cite}
\usepackage{amsmath,amssymb,amsfonts}
\usepackage{import}
\usepackage{subcaption}
\usepackage{algorithmic}
\usepackage{graphicx}
\usepackage{textcomp}
\usepackage{xcolor}
\usepackage{balance}
\usepackage{algorithm} 

\usepackage{algorithmic}
\usepackage{booktabs}
\usepackage{multirow}
\usepackage{comment}
\def\BibTeX{{\rm B\kern-.05em{\sc i\kern-.025em b}\kern-.08em
    T\kern-.1667em\lower.7ex\hbox{E}\kern-.125emX}}
\def\bumpup{\vspace{-0.5cm}}
\def\bumpdown{\vspace{0.3cm}}
\begin{document}

\title{SR4-Fit: An Interpretable and Informative Classification Algorithm Applied to Prediction of U.S. House of Representatives Elections}

\author{
    \IEEEauthorblockN{Shyam Sundar Murali Krishnan}
    \IEEEauthorblockA{
        \textit{School of Computer Science}\\
        \textit{Gallogly College of Engineering}\\
        \textit{University of Oklahoma}\\
        Norman, Oklahoma, USA \\
        shyamkrishnan@ou.edu
    }
\and
    \IEEEauthorblockN{Dean Frederick Hougen}
    \IEEEauthorblockA{
        \textit{School of Computer Science}\\
        \textit{Gallogly College of Engineering}\\    
        \textit{University of Oklahoma}\\
        Norman, Oklahoma, USA \\
        hougen@ou.edu}
}

\begin{comment}
\author{\IEEEauthorblockN{Author 1}
\IEEEauthorblockA{
\textit{Institution 1}\\
Location 1\\
Email Address 1}
\and
\IEEEauthorblockN{Author 2}
\IEEEauthorblockA{
\textit{Institution 1}\\
Location 1\\
Email Address 2}
}
\end{comment}

\maketitle

\begin{abstract} 

The growth of machine learning demands interpretable models for critical applications, yet most high-performing models are ``black-box'' systems that obscure input-output relationships, while traditional rule-based algorithms like RuleFit suffer from a lack of predictive power and instability despite their simplicity. This motivated our development of Sparse Relaxed Regularized Regression Rule-Fit (SR4-Fit), a novel interpretable classification algorithm that addresses these limitations while maintaining superior classification performance. Using demographic characteristics of U.S. congressional districts from the Census Bureau's American Community Survey, we demonstrate that SR4-Fit can predict House election party outcomes with unprecedented accuracy and interpretability. Our results show that while the majority party remains the strongest predictor, SR4-Fit has revealed intrinsic combinations of demographic factors that affect prediction outcomes that were unable to be interpreted in black-box algorithms such as random forests. The SR4-Fit algorithm surpasses both black-box models and existing interpretable rule-based algorithms such as RuleFit with respect to accuracy, simplicity, and robustness, generating stable and interpretable rule sets while maintaining superior predictive performance, thus addressing the traditional trade-off between model interpretability and predictive capability in electoral forecasting. To further validate SR4-Fit's performance, we also apply it to six additional publicly available classification datasets, like the breast cancer, Ecoli, page blocks, Pima Indians, vehicle, and yeast datasets,  and find similar results.

\end{abstract}


\section{Introduction}
\label{sec:Introduction}
Modern machine learning (ML) techniques, such as random forests \cite{breiman2001random}, Support Vector Machines (SVMs) \cite{hearst1998support}, and neural networks \cite{mcculloch1943logical}, which have shown strong results in a variety of applications, are regarded as ``black boxes'' because it is difficult to understand the sophisticated mappings these methods learn from inputs to outputs. Because these models contain a large number of operations, they become very difficult for humans to understand, even though they contribute to more accurate forecasts. This human incomprehensibility of the models could cause issues in applications where important choices are at risk (e.g.,~\cite{caruana2015intelligible}). Due to this issue, there is growing emphasis on improving model interpretability, which gives rise to the idea of \emph{interpretable machine learning}. If the model aids in comprehending the general behavior of the data, it can be deemed \emph{interpretable} and assist people in comprehending and having faith in the models~\cite{molnar2020interpretable}.

One important domain where interpretability is critical is in the context of U.S. House of Representatives elections. In a democratic system like that of the United States, understanding how representatives are elected is essential to ensuring fair political representation. Election outcomes are influenced not only by how individuals vote but also by how voting districts are drawn. The demographic composition of a district, including factors such as race, income, education, and age, plays a crucial role in shaping voting behavior and, in turn, determining party representation \cite{arrington2010affirmative}. By quantifying correlations between demographic variables and party alignment, researchers can better evaluate existing electoral trends and uncover new patterns. Moreover, these analyses allow for the prediction and evaluation of \emph{district security}, which refers to the likelihood that a party will retain control over a district in future elections. This metric is especially valuable during redistricting, where it can help assess whether districts are being drawn fairly or manipulated to favor a particular party, a practice known as \emph{gerrymandering} \cite{friedman2009rising}. Ultimately, predictive and interpretable models enable a deeper understanding of political dynamics and support the design of electoral districts that are both representative and equitable.

Previous research has shown that it is possible to predict U.S. congressional district elections using machine learning models such as random forests and SVMs, which have demonstrated strong predictive performance and have helped bridge the gap between computational tools and the practical demands of electoral forecasting \cite{richardson2020districts}. However, a critical question remains: Are these predictive models truly reliable, and how can we understand the underlying patterns and behaviors to validate their outputs? While models like random forests can generate decision rules that reflect data patterns, their internal structure is often dense and convoluted, leading to an overwhelming number of complex rules that are difficult to interpret. This complexity makes it challenging to extract meaningful insights or assess whether the model’s decisions align with real-world political dynamics. As a result, relying solely on predictive accuracy is insufficient, especially in high-stakes applications like election forecasting, where transparency and trust are essential. This highlights the growing need for interpretable machine learning techniques that prioritize models whose internal logic is comprehensible and whose predictions can be validated by domain experts. Interpretability in machine learning can be achieved either through post-hoc methods or intrinsic approaches \cite{lipton2018mythos}.

\emph{Post-hoc interpretability} involves explaining complex models after training using tools like SHapley Additive exPlanations (SHAP) \cite{lundberg2017unified} or  Local Interpretable Model-agnostic Explanations (LIME) \cite{ribeiro2016should}, but these explanations are often approximations and can be misleading.

In contrast, \emph{intrinsic interpretability} refers to models that are transparent by design, such as decision trees or rule-based models, where the decision-making process is naturally understandable. Intrinsic models are especially valuable in sensitive domains because they provide faithful, robust, and easily verifiable insights without relying on surrogate explanations like the post-hoc approaches \cite{rudin2019stop}. Rule-based models are widely used for interpretable classification because they generate transparent rules that clearly explain how decisions are made. However, many traditional rule-based models face key limitations: some are restricted to binary classification \cite{obregon2023rulecosi+}, while others prioritize simplicity so heavily that they sacrifice predictive accuracy. This highlights the classic interpretability–accuracy trade-off, where improving one often comes at the cost of the other \cite{murdoch2019definitions}.

Models like RuleFit attempt to strike a balance by combining rule extraction with linear modeling to enhance both interpretability and performance. Still, they may underperform on complex real-world data \cite{yang2017scalable}, making it difficult to trust their predictions despite their readable rules. Since meaningful interpretability relies not just on rule simplicity but also on predictive reliability, models that achieve both are especially valuable, enabling a deeper understanding of feature influence in sensitive tasks like election prediction and demographic-based classification. 

This led to our development of a rule-based model called \emph{Sparse Relaxed Regularized Regression Rule Fit} (SR4-Fit), which builds on the interpretability framework of the RuleFit algorithm \cite{friedman2008predictive} and incorporates the sparse optimization capabilities of Sparse Relaxed Regularized Regression (SR3) \cite{zheng2018unified}. Results from our experiments show that SR4-Fit consistently performs better than both RuleFit and random forests in terms of interpretability, while matching or slightly exceeding their predictive accuracy—effectively bridging the gap between transparency and performance. Both RuleFit and random forests demonstrate noticeable variability across trials, making them less consistent and less reliable for stable predictions. In contrast, SR4-Fit provides more consistent outcomes and clearer insights into the model’s decision-making process. SR4-Fit generates concise, interpretable rules that highlight meaningful demographic patterns—such as the influence of age groups, educational attainment, and racial composition on voting outcomes. Furthermore, SR4-Fit is not limited to binary classification and performs effectively on multi-class datasets, enhancing its applicability to a broader range of problems, suggesting that it offers a practical and well-balanced approach, combining predictive reliability, model stability, and interpretability, making it a valuable tool for electoral forecasting and understanding party alignment through demographic data, as well as being broadly applicable to other important classification problems.

The remainder of the paper is structured as follows. Section~\ref{sec:SR4-Fit Implementation} introduces the SR4-Fit classification algorithm, Section~\ref{sec:Results} outlines the datasets and experimental setup, Section~\ref{sec:Discussion} presents results and discussion, and Section~\ref{sec:Conclusion} presents conclusions and future work.

\section{SR4-Fit Implementation}
\label{sec:SR4-Fit Implementation}

The SR4-Fit algorithm is developed based on the interpretability approach of the RuleFit algorithm with the sparse optimization capabilities of SR3. SR4-Fit extracts decision rules from decision trees to capture non-linear patterns similar to RuleFit, and then applies SR3 optimization to fit a sparse linear model over both rules and raw features. To adapt this setup for classification tasks, the SR3 objective is modified by replacing the squared loss with logistic loss, allowing the model to predict class probabilities rather than continuous outputs. This results in a sparse, interpretable classification model that leverages rule-based logic and SR3's structured regularization to balance accuracy with explainability.\footnote{Python code available at: https://github.com/Shyamyohaan96/SR4\_IML}

\subsection{Rule Extraction}
\label{Rule Extraction}

The SR4-Fit algorithm starts by extracting interpretable decision rules from a collection of decision trees, which are trained on the input data. For each class in the dataset, a binary one-versus-rest formulation \cite{rifkin2004defense} is used, where the forest is trained to distinguish the class of interest from all others. From every tree in the forest, rules are extracted from each path from the root to internal or terminal nodes. Each path defines a rule composed of logical conditions on the input features. These rules are encoded as binary indicators, returning 1 if the rule evaluates to true for a particular input vector and 0 otherwise. For example, the rule
\begin{equation*}
X_1 \leq 2.7 \ \text{and} \ X_3 > 1.4
\label{eq:rul}
\end{equation*}
asserts that input feature $X_1$ is less than 2.7 and input feature $X_3$ is greater than 1.4. If this is true, the rule evaluates to 1; otherwise, it evaluates to 0. Because rules are collected from each tree throughout, the resulting set captures diverse and robust patterns throughout the input space. To ensure interpretability, the number of rules retained is limited by the maximum rules $r_{\text{max}}$ parameter.

\subsection{Feature Construction and Optimization Objective}
\label{Feature Construction and Optimization Objective}

Following rule extraction, SR4-Fit builds an extended feature matrix by combining the original input features with the binary rule indicators. For each data point, the rule evaluations are concatenated with the original feature values to form a comprehensive representation $Z = [X \;|\; R]$ where $X \in \mathbb{R}^{n \times d}$ and $R \in \{0, 1\}^{n \times m}$. This expanded feature matrix allows the model to utilize both raw features and data-driven patterns captured by the trees. To learn a sparse and accurate classifier, the algorithm minimizes the loss function, which consists of two parameters $\beta$, the model's predictive weight vector, and $w$, the vector enforcing sparsity:
\begin{equation*}
L(\beta, w) = \sum_{i=1}^{n} \log\left(1 + \exp\left(-y_i Z_i^\top \beta\right)\right) + \lambda \|w\|_1 + \frac{\kappa}{2} \|\beta - w\|_2^2.
\label{eq:loss_function}
\end{equation*}
In this objective function, $\lambda$ controls the sparsity and $\kappa$ controls how closely $\beta$ must follow $w$. The algorithm minimizes this objective function by alternating between the two update steps. First, with $w$ fixed, $\beta$ is updated using gradient-based optimization on the logistic loss for fitting accuracy and quadratic regularization for enforcing $\beta \approx w$. Second, with $\beta$ fixed, $w$ is updated via soft-thresholding, which is key for achieving sparsity \cite{tibshirani1996regression}, using

\begin{equation*}
w_j = \operatorname{sign}(\beta_j) \cdot \max\left(|\beta_j| - \frac{\lambda}{\kappa},\, 0\right)
\label{eq:soft_threshold}
\end{equation*}
where
\begin{equation*}
\operatorname{sign}(z) =
\begin{cases}
+1, & \text{if } z > 0 \\
\;\;0, & \text{if } z = 0 \\
-1, & \text{if } z < 0.
\end{cases}
\label{eq:sign}
\end{equation*}

This iterative process continues until convergence, producing both a predictive and sparse representation of the input features.

\subsection{Model Pruning and Rule Selection}
\label{Model Pruning and Rule selection}

Once the optimization is done, the algorithm identifies the most relevant features and rules by examining the optimized sparse vector $w$. Any component $w_j = 0$ indicates that the corresponding rule or feature does not contribute to the prediction and is excluded from the final model. This pruning removes redundancy and focuses on a small number of predictors that are informative to the model's effectiveness. Since the objective function separates the predictive quality, which is handled by $\beta$, from the sparsity enforcement, which is handled by $w$, it enforces an effective rule selection process compared to methods that enforce sparsity directly on the predictive weights.

\subsection{Classification and Probability Estimation}
\label{Classification and Probability Estimation}

In the final step, SR4-Fit uses the learned coefficients $\beta$ to make predictions. For the new input sample, the model evaluates the selected rules and features and computes their linear combination using
\begin{equation*}
s(x) = \beta_0 + \sum_{j=1}^{d+m} \beta_j Z_j(x).
\label{eq:prediction_function}
\end{equation*}

Finally, SR4-Fit applies the sigmoid function to produce the predicted probability using
\begin{equation*}
\hat{y}(x) = \sigma(s(x)) = \frac{1}{1 + \exp(-s(x))}.
\label{eq:sigmoid_output}
\end{equation*}

This probability reflects the model's confidence in assigning the input to the positive class. For multiclass problems, the algorithm fits one SR4-Fit classifier per class using a one-vs-rest strategy, and the predicted class corresponds to the highest predicted probability.

\begin{algorithm}
    \caption{SR4-Fit Classification Algorithm}
    \begin{algorithmic}[1]
        \STATE \textbf{Input:} Dataset $(X, y)$, parameters $\lambda$, $\kappa$ and $r_{\text{max}}$.
        \STATE \textbf{Output:} Predictive weights $\beta$.
        \FOR{each class $c = 0, 1, \ldots, C$}
            \STATE Train a random forest using one-vs-rest labeling for class $c$ \cite{breiman2001random}.
            \STATE Extract decision rules from each path in each tree until $r_{\text{max}}$ \cite{friedman2001greedy}.
            \STATE Encode rules as binary indicators $R \in \{0,1\}^{n \times m}$ \cite{cohen1995fast}.
            \STATE Form an extended feature matrix $Z = [X \;|\; R]$.
            \STATE Initialize $\beta$, $w$.
            \WHILE{not converged}
                \STATE Minimize $L(\beta, w) = \newline
                \hspace*{1em}\displaystyle\sum_{i=1}^n \log(1 + \exp(-y_i Z_i^\top \beta)) + \lambda \|w\|_1 + \frac{\kappa}{2}\|\beta - w\|_2^2$
                \STATE Update $w$,  $w_j = \operatorname{sign}(\beta_j) \cdot \max\left(|\beta_j| - \frac{\lambda}{\kappa},\, 0\right)$
            \ENDWHILE
            \STATE Prune rules where $w_j = 0$
        \ENDFOR
        \STATE \textbf{Prediction:}
        \STATE Compute $s(x) = \beta_0 + \displaystyle\sum_{j=1}^{d+m} \beta_j Z_j(x)$
        \STATE Compute predicted probability $\hat{y}(x) = \frac{1}{1 + \exp(-s(x))}$
    \end{algorithmic}
\end{algorithm}

\section{Experiments}
\label{sec:Results}
\subsection{Dataset Background}

The demographic data, which was used in a previous study covering U.S. House of Representatives elections from 2006 to 2013 for training and 2014 to 2016 for testing \cite{census_s0601_2006_2016} has been used for these experiments. The election results were obtained from the MIT Election Data and Science Lab’s “U.S. House 1976–2018” dataset \cite{medsl_us_house_1976_2018}, which provides district-level outcomes. Of the 2,610 seats contested during this period, seven (0.2\%) were won by candidates who were neither Democrats nor Republicans. For the experiments, four datasets were used based on the demographic characteristics included. The \emph{minimum dataset} consists of median age, male percentage, white percentage, and bachelor’s degree attainment. The \emph{standard dataset} adds breakdowns of age, race, income, and education categories to the minimum set. The \emph{expanded dataset} further includes language spoken at home, marital status, and poverty indicators. The \emph{previous dataset} supplements the standard attributes with information about the party that won the district in the prior election. In addition, all four datasets include the percentages of both registered Democrats and Republicans.

To demonstrate the broad applicability of the SR4-Fit algorithm in classification settings, six publicly available datasets were also selected for experimentation. These include both binary and multiclass classification tasks, as well as imbalanced datasets, ensuring comprehensive evaluation. The Wisconsin Breast Cancer dataset \cite{breast_cancer_wisconsin} provides diagnostic features from digitized images of breast masses to classify tumors as benign or malignant. The Ecoli dataset \cite{ecoli_39} is comprised of protein localization sites across different cellular compartments. The Page Blocks dataset \cite{page_blocks_classification_78} includes layout and geometric features of document blocks from scanned pages for classifying each block’s type. The Pima Indians Diabetes dataset \cite{uci_pima_diabetes} contains medical and demographic attributes of female Pima Indian individuals to predict the onset of type 2 diabetes. The Vehicle Silhouette dataset \cite{statlog_(vehicle_silhouettes)} involves shape descriptors extracted from vehicle silhouettes to classify them into categories based on geometric properties. Lastly, the Yeast dataset \cite{yeast_110} includes sequence-based features of yeast proteins to predict their cellular localization.

\subsection{Experimentation}

Previous research on U.S. demographic data had shown that Random Forest and Support Vector Machine (SVM) models are strong predictors. Therefore, both types of models were selected for experimentation and comparison with SR4-Fit to evaluate performance on demographic data, enabling a direct comparison between black-box models and the interpretable SR4-Fit rule-based model. Additionally, SR4-Fit was compared with the traditional RuleFit algorithm, as SR4-Fit was designed with a similar interpretability objective, ensuring a fair evaluation setting. The hyperparameters ($r_{\text{max}}$, $\lambda$ , $\kappa$) were selected via grid search \cite{pedregosa2011scikit} across a range of values. Each setting was evaluated over multiple trials, and the combination yielding higher accuracy was chosen.

All experiments on the U.S. demographic datasets were conducted over 30 independent trials for each model to provide for statistically meaningful results. The predictive performance was assessed using standard classification metrics, including accuracy, precision, recall \cite{swets1969effectiveness}, and F1-score \cite{van1979information}, to provide a quantitative evaluation. The Dice–Sorensen index \cite{chao2006abundance} was also employed to analyze the robustness of the models.

To quantify the interpretability of the model, the \emph{Interpretability Score} (IPS) is calculated based on the equally weighted combination of the accuracy of the model, the consistency or robustness of the model (Dice-Sorensen), the simplicity of the model (number of rules), and the rule complexity (number of components in a rule) \cite{margot2021new}.

To facilitate comparative analysis, violin plots, line plots, and box plots were used \cite{tukey1977exploratory}. These visualizations help to illustrate the variability, central tendency, and distribution of each model's performance in aggregate and across trials, making it easier to assess both consistency and predictive strength \cite{suh2023metrics}. The same evaluation protocol, including metrics and visual analysis, was applied to standard benchmark datasets across 10 trials per model.

\section{Results and Discussion}
\label{sec:Discussion}

This section covers the performance of all four ML approaches for U.S. election forecasting (Subsection~\ref{ss:results-election}), discussion of the interpretability of SR4-Fit's rules on that dataset (Subsection~\ref{ss:SR4-Fit-interpretation}), and the performance of all four ML approaches to more commonly used public datasets (Subsection~\ref{ss:result-public}).  

\subsection{Performance on Election Forecasting Data}
\label{ss:results-election}

\begin{figure}[htbp]
    \centering
    %\vspace{-0.6cm}
    \begin{subfigure}[b]{\linewidth}
        \centering
        \includegraphics[width=\linewidth]{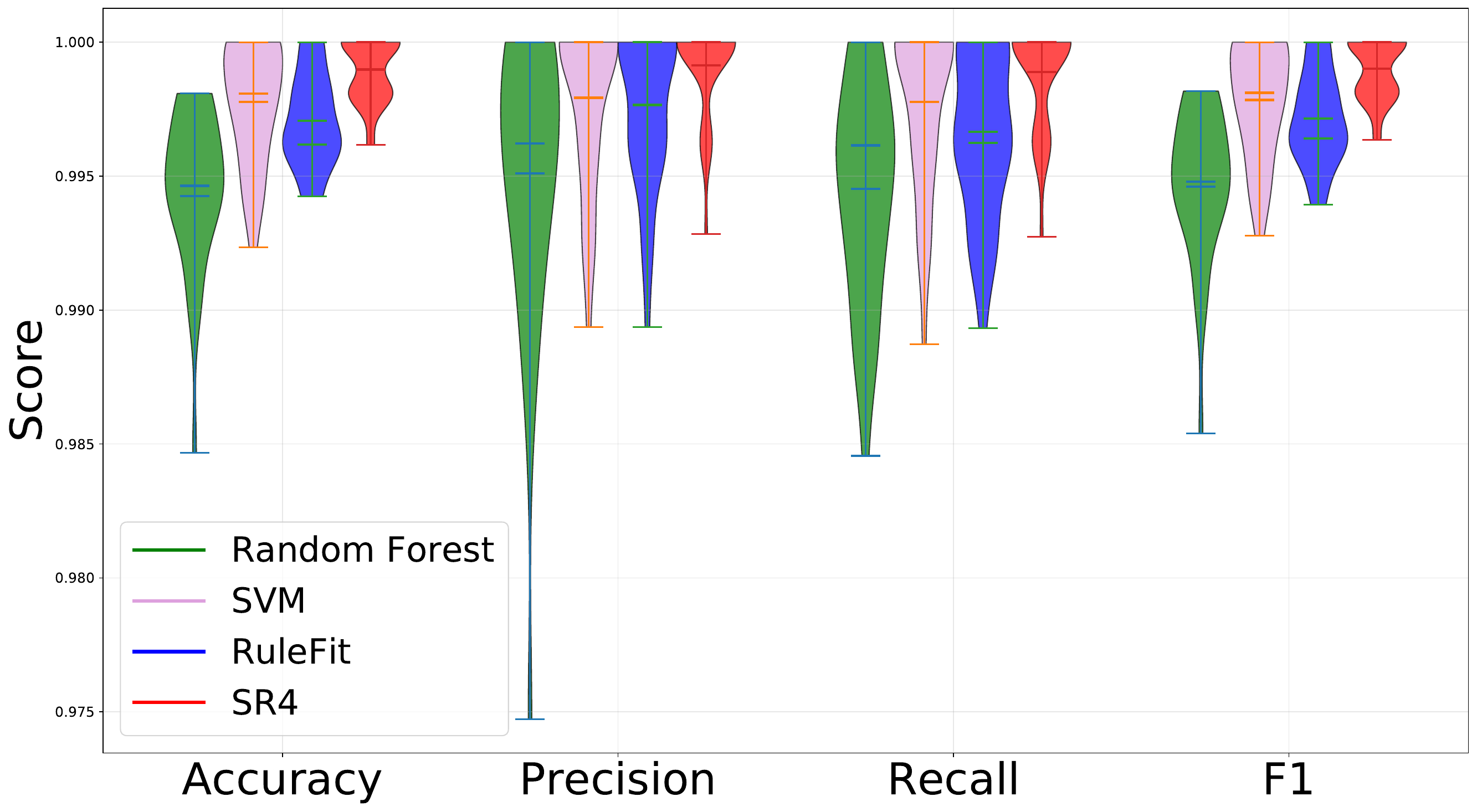}
        \bumpup
        \caption{Minimum Data}
        \bumpdown
        \label{fig:violin_min}
    \end{subfigure}

    \begin{subfigure}[b]{\linewidth}
        \centering
        \includegraphics[width=\linewidth]{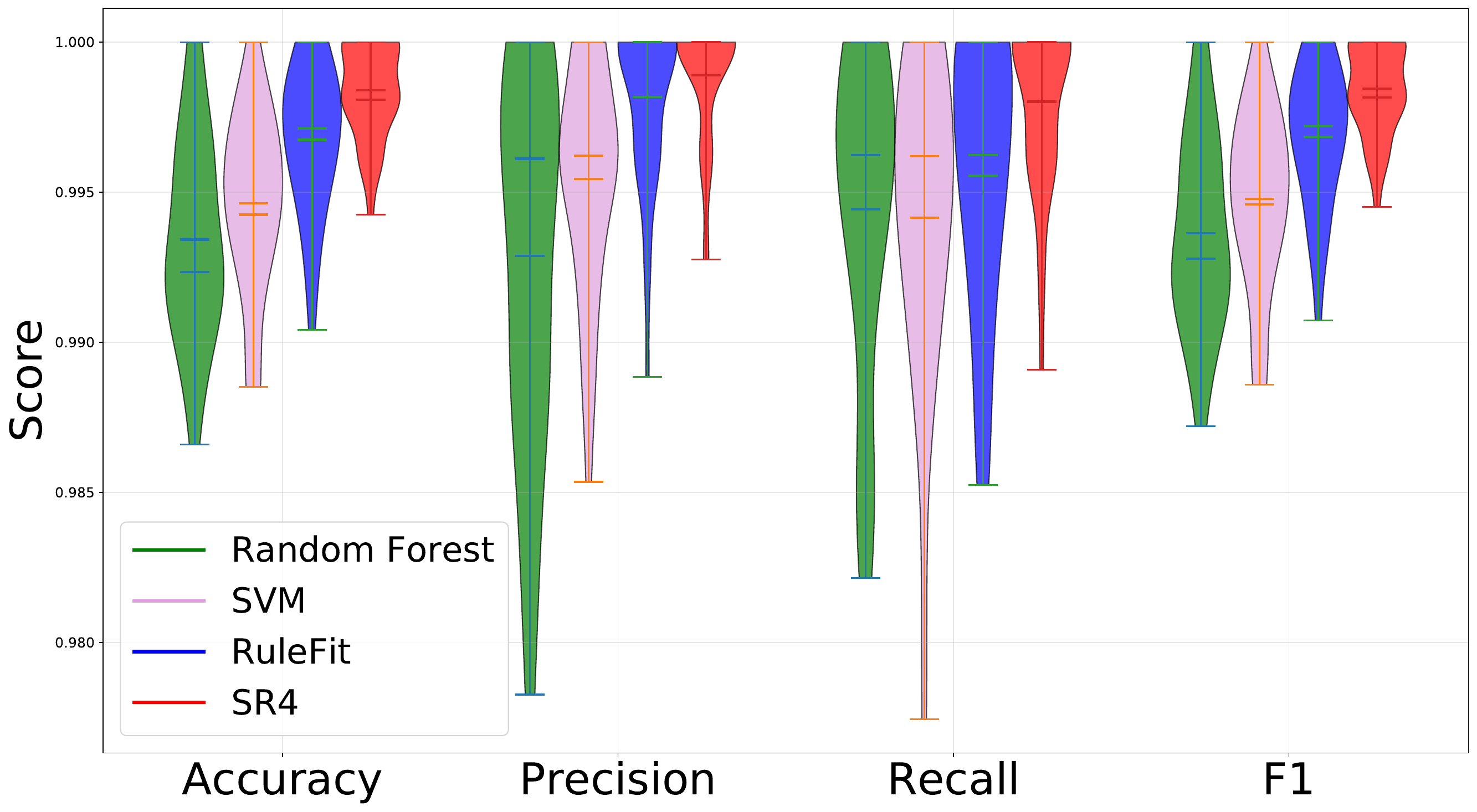}
        \bumpup
        \caption{Standard Data}
        \bumpdown
        \label{fig:violin_std}
    \end{subfigure}

    \begin{subfigure}[b]{\linewidth}
        \centering
        \includegraphics[width=\linewidth]{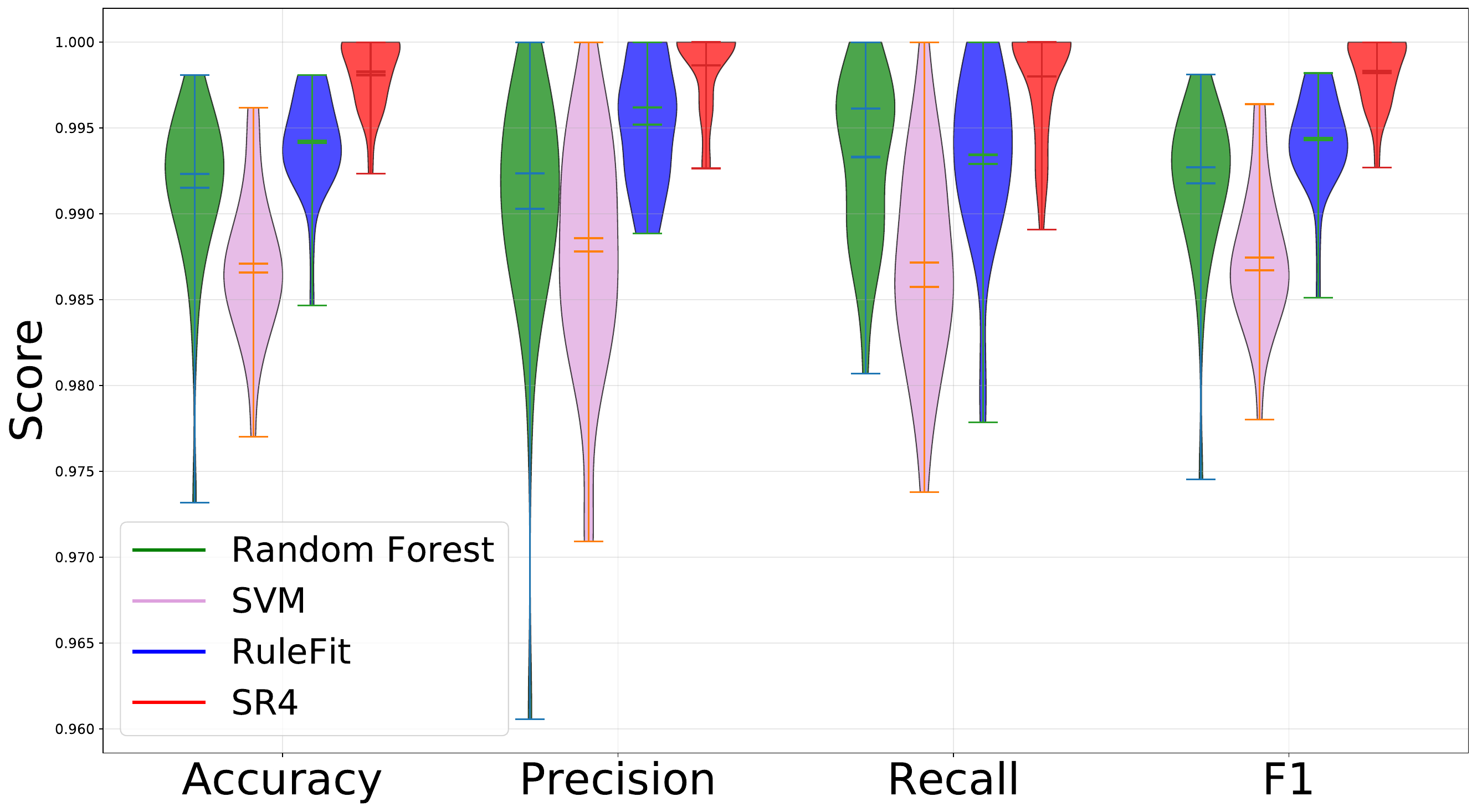}
        \bumpup
        \caption{Expanded Data}
        \bumpdown
        \label{fig:violin_exp}
    \end{subfigure}

    \begin{subfigure}[b]{\linewidth}
        \centering
        \includegraphics[width=\linewidth]{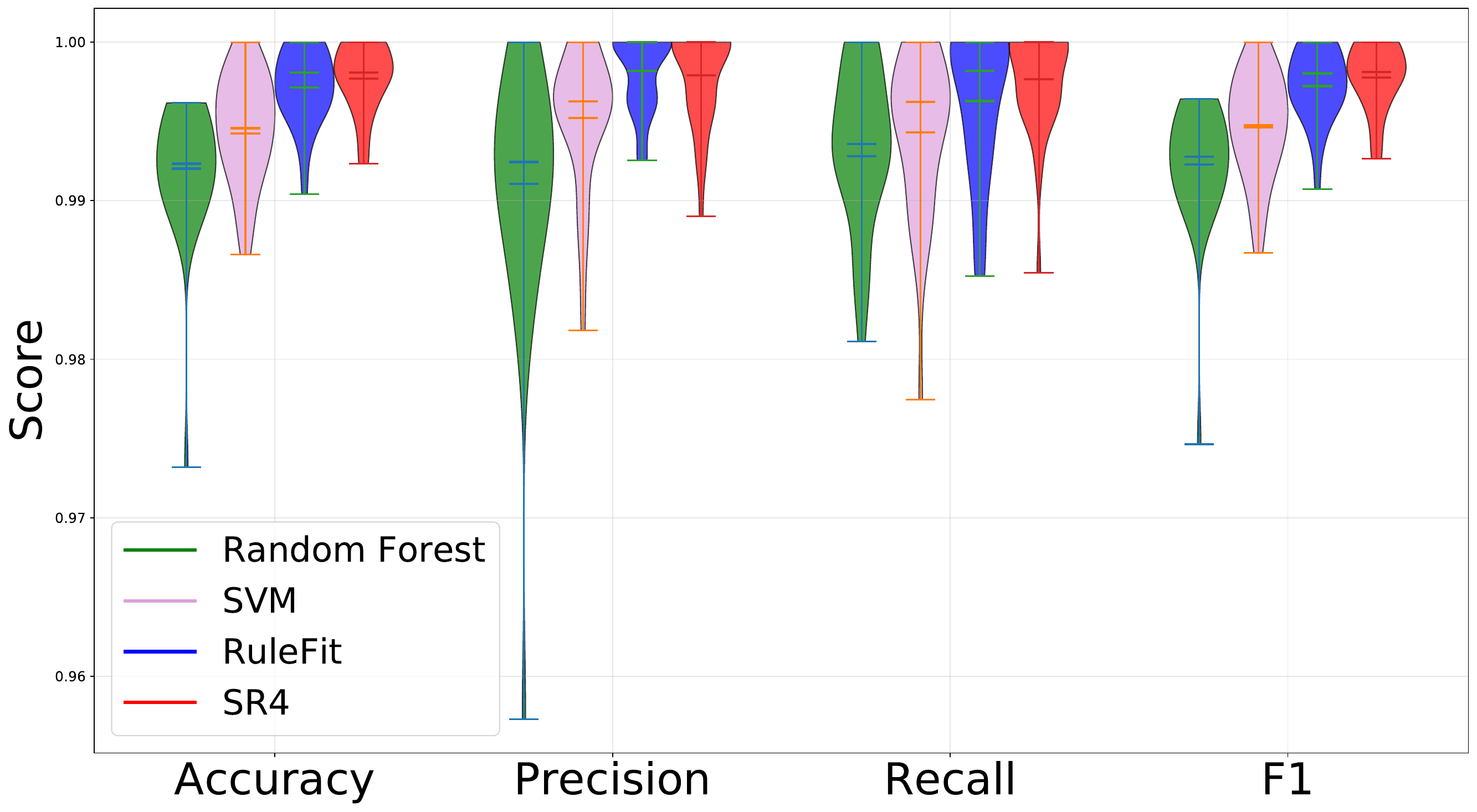}
        \bumpup
        \caption{Previous Data}
        \bumpdown
        \label{fig:violin_pre}
    \end{subfigure}

    \vspace{-0.2cm}
    \caption{Violin plot comparison of all models across different datasets for different prediction metrics.}
    \label{fig:violin_all_models}
\end{figure}

The findings from the experimental evaluation across the minimum, standard, expanded, and previous datasets are summarized in figures and tables emphasizing predictive accuracy, precision, recall, F1-score, the t-tests and p-value tests between RuleFit and SR4-Fit, model stability, and interpretability. Violin plots for accuracy, precision, recall, and F1 score are found in Figure~\ref{fig:violin_all_models} and violin plots for IPS are found in Figure~\ref{fig:ips_violin}. All other plots are found in the supplementary material.

\begin{table}[htbp]
\setlength{\tabcolsep}{2pt}
\centering
\resizebox{\linewidth}{!}{  % scales table to column width
\begin{tabular}{lcccc}
\toprule
\textbf{Dataset} & \textbf{Random Forest} & \textbf{SVM} & \textbf{RuleFit} & \textbf{SR4-Fit}\\
\midrule
Minimum & 0.9946$\pm$0.0029 & 0.9978$\pm$0.0022 & 0.9971$\pm$0.0018 & \textbf{0.9990$\pm$0.0012} \\
Standard & 0.9934$\pm$0.0033 & 0.9946$\pm$0.0029 & 0.9967$\pm$0.0024 & \textbf{0.9984$\pm$0.0016} \\ 
Expanded & 0.9915$\pm$0.0047 & 0.9871$\pm$0.0041 & 0.9941$\pm$0.0029 & \textbf{0.9983$\pm$0.0020} \\ 
Previous & 0.9920$\pm$0.0043 & 0.9946$\pm$0.0034 & 0.9971$\pm$0.0024 & \textbf{0.9977$\pm$0.0022} \\ 
\bottomrule
\end{tabular}
}
%\vspace{0.5em}
\caption{Average accuracy $\pm$ standard deviation over 30 trials for different demographic datasets and models. \textbf{Bold} indicates best performance.}
\label{tab:Accuracy}
\end{table}

\begin{table}[htbp]
\centering
\setlength{\tabcolsep}{2pt}
\resizebox{\linewidth}{!}{
\begin{tabular}{lcccc}
\toprule
\textbf{Dataset} & \textbf{Random Forest} & \textbf{SVM} & \textbf{RuleFit} & \textbf{SR4-Fit} \\
\midrule
Minimum  & 0.9951$\pm$0.0055 & 0.9979$\pm$0.0031 & 0.9977$\pm$0.0029 & \textbf{0.9991$\pm$0.0018} \\
Standard & 0.9929$\pm$0.0062 & 0.9954$\pm$0.0038 & 0.9982$\pm$0.0028 & \textbf{0.9989$\pm$0.0021} \\
Expanded & 0.9903$\pm$0.0072 & 0.9878$\pm$0.0068 & 0.9952$\pm$0.0036 & \textbf{0.9987$\pm$0.0024} \\
Previous & 0.9911$\pm$0.0080 & 0.9952$\pm$0.0043 & \textbf{0.9982$\pm$0.0024} & 0.9979$\pm$0.0029 \\
\bottomrule
\end{tabular}
}
\caption{Average precision $\pm$ standard deviation over 30 trials for different demographic datasets and models. \textbf{Bold} indicates best performance.}
\label{tab:precision}
\end{table}

\begin{table}[htbp]
\centering
\setlength{\tabcolsep}{2pt}
\resizebox{\linewidth}{!}{
\begin{tabular}{lcccc}
\toprule
\textbf{Dataset} & \textbf{Random Forest} & \textbf{SVM} & \textbf{RuleFit} & \textbf{SR4-Fit} \\
\midrule
Minimum  & 0.9945$\pm$0.0041 & 0.9978$\pm$0.0031 & 0.9966$\pm$0.0031 & \textbf{0.9989$\pm$0.0019} \\
Standard & 0.9944$\pm$0.0053 & 0.9942$\pm$0.0053 & 0.9955$\pm$0.0045 & \textbf{0.9980$\pm$0.0028} \\
Expanded & 0.9933$\pm$0.0049 & 0.9872$\pm$0.0059 & 0.9934$\pm$0.0051 & \textbf{0.9980$\pm$0.0032} \\
Previous & 0.9936$\pm$0.0049 & 0.9943$\pm$0.0051 & 0.9963$\pm$0.0046 & \textbf{0.9977$\pm$0.0032} \\
\bottomrule
\end{tabular}
}
\caption{Average recall $\pm$ standard deviation over 30 trials for different demographic datasets and models. \textbf{Bold} indicates best performance.}
\label{tab:recall}
\end{table}

\begin{table}[htbp]
\centering
\setlength{\tabcolsep}{2pt}
\resizebox{\linewidth}{!}{
\begin{tabular}{lcccc}
\toprule
\textbf{Dataset} & \textbf{Random Forest} & \textbf{SVM} & \textbf{RuleFit} & \textbf{SR4-Fit} \\
\midrule
Minimum  & 0.9948$\pm$0.0028 & 0.9978$\pm$0.0021 & 0.9971$\pm$0.0017 & \textbf{0.9990$\pm$0.0011} \\
Standard & 0.9936$\pm$0.0032 & 0.9948$\pm$0.0029 & 0.9968$\pm$0.0023 & \textbf{0.9985$\pm$0.0015} \\
Expanded & 0.9918$\pm$0.0045 & 0.9875$\pm$0.0040 & 0.9943$\pm$0.0028 & \textbf{0.9983$\pm$0.0019} \\
Previous & 0.9923$\pm$0.0040 & 0.9947$\pm$0.0033 & 0.9972$\pm$0.0023 & \textbf{0.9978$\pm$0.0021} \\
\bottomrule
\end{tabular}
}
\caption{Average F1-score $\pm$ standard deviation over 30 trials for different demographic datasets and models. \textbf{Bold} indicates best performance.}
\label{tab:F1}
\end{table}

\begin{table}[htbp]
\setlength{\tabcolsep}{1.5pt}
\centering
\resizebox{\linewidth}{!}{
\begin{tabular}{lcccccccc}
\toprule
\multirow{2}{*}{\textbf{Dataset}} & \multicolumn{2}{c}{\textbf{Accuracy}} & \multicolumn{2}{c}{\textbf{Precision}} & \multicolumn{2}{c}{\textbf{Recall}} & \multicolumn{2}{c}{\textbf{F1 Score}} \\
\cmidrule(lr){2-3} \cmidrule(lr){4-5} \cmidrule(lr){6-7} \cmidrule(lr){8-9}
& \textbf{t-test} & \textbf{p-value} & \textbf{t-test} & \textbf{p-value} & \textbf{t-test} & \textbf{p-value} & \textbf{t-test} & \textbf{p-value} \\
\midrule
Minimum  & \textbf{-4.8534} & \textbf{0.0000} & \textbf{-2.3397} & \textbf{0.0228} & \textbf{-3.3390} & \textbf{0.0015} & \textbf{-4.8855} & \textbf{0.0000} \\
Standard & \textbf{-3.0894} & \textbf{0.0031} & -1.1125 & 0.2705 & \textbf{-2.5069} & \textbf{0.0150} & \textbf{-3.1019} & \textbf{0.0030} \\
Expanded & \textbf{-6.3727} & \textbf{0.0000} & \textbf{-4.3444} & \textbf{0.0001} & \textbf{-4.1098} & \textbf{0.0001} & \textbf{-6.3983} & \textbf{0.0000} \\
Previous & -0.9638 & 0.3392 & 0.3917 & 0.6967 & -1.3190 & 0.1923 & -1.9222 & 0.0595 \\
\bottomrule
\end{tabular}
}
\caption{T-test and p-value comparison between RuleFit and SR4-Fit over 30 trials across demographic datasets. Statistically significant differences ($p < 0.05$) are shown in \textbf{bold}.}
\label{tab:ttests_split}
\end{table}

\begin{table}[htbp]
\centering
\begin{tabular}{lccc}
\toprule
\textbf{Dataset} & \textbf{Random Forest} & \textbf{RuleFit} & \textbf{SR4-Fit}\\
\midrule
Minimum & 0.0512$\pm$0.0110 & \textbf{0.4776$\pm$0.01615} & \textbf{0.4776$\pm$0.01615} \\
Standard & 0.0313$\pm$0.0089 & \textbf{0.7814$\pm$0.0111} &0.7370$\pm$0.0110 \\ 
Expanded & 0.0385$\pm$0.0109 & \textbf{0.8540$\pm$0.0079} & \textbf{0.8540$\pm$0.0079} \\ 
Previous & 0.0606$\pm$0.0103 & 0.7437$\pm$0.0108 & \textbf{0.7873$\pm$0.0108} \\ 
\bottomrule
\end{tabular}
%\vspace{0.5em}
\caption{Average stability $\pm$ standard deviation (Dice-Sorensen index) run across 30 trials for different demographic datasets across different model combinations (rule-based only). \textbf{Bold} indicates best performance.}
\label{tab:Stability}

\end{table}

\begin{table}[htbp]
\centering
\begin{tabular}{lccc}
\toprule
\textbf{Dataset} & \textbf{Random Forest} & \textbf{RuleFit} & \textbf{SR4-Fit}\\
\midrule
Minimum & 0.5075$\pm$0.1507 & 0.5821$\pm$0.1555 & \textbf{0.6432$\pm$0.1706} \\
Standard & 0.5019$\pm$0.1554 & 0.6446$\pm$0.2384 &\textbf{0.6594$\pm$0.1673} \\ 
Expanded & 0.5022$\pm$0.1313 & 0.6562$\pm$0.2289 & \textbf{0.6738$\pm$0.2101} \\ 
Previous & 0.5239$\pm$0.1244 & 0.6542$\pm$0.1732 & \textbf{0.6548$\pm$0.2293} \\ 
\bottomrule
\end{tabular}

%\vspace{0.5em}
\caption{Average interpretability score $\pm$ standard deviation run across 30 trials for different demographic datasets across different model combinations (rule-based only). \textbf{Bold} indicates best performance.}
\label{tab:ips}
\end{table}

\begin{figure}[htbp]
   % \centering
    \includegraphics[width=1.0\linewidth]{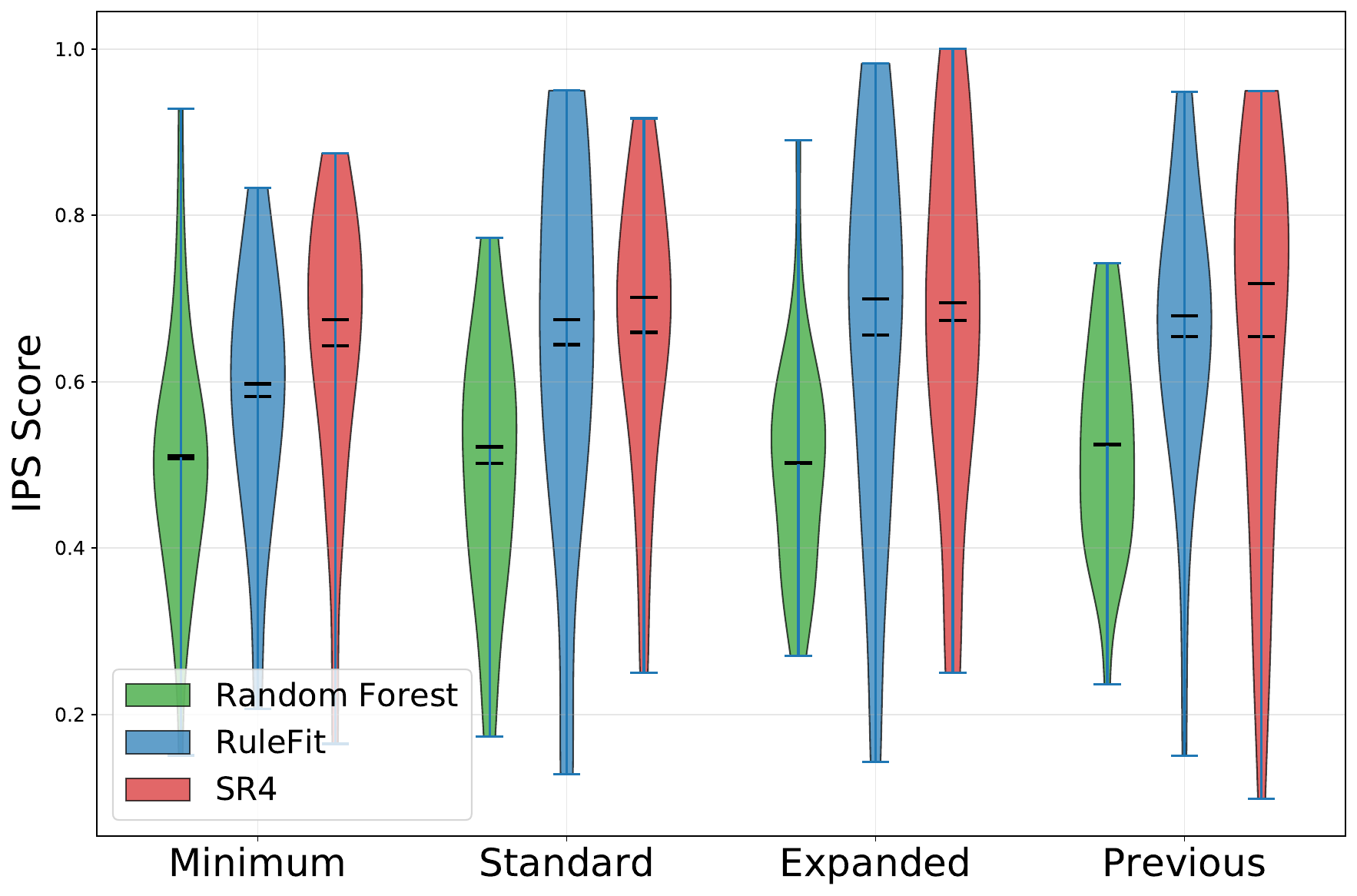}
    \caption{Violin plot comparison of the interpretability score of all rule-based models across different demographic datasets}
    \label{fig:ips_violin}
\end{figure}

In the minimum dataset, SR4-Fit achieved the highest accuracy, precision, recall, and F1 score with minimal variance, as shown in Table~\ref{tab:Accuracy} through Table~\ref{tab:F1}, and reflected in the narrow violin distributions in Figure~\ref{fig:violin_min}. This is further supported by the narrow box plot distributions and flat line plots across trials (see Supplementary Material, Section A), indicating outstanding stability and generalization even with sparse features. RuleFit followed closely in accuracy but exhibited slightly higher variability in recall and F1-score. Notably, both rule-based interpretable algorithms, SR4-Fit and RuleFit, achieved high rule stability, significantly outperforming random forest, which exhibited erratic behavior across all plots, as shown in Table~\ref{tab:Stability}.

In the standard dataset, SR4-Fit continued to lead in accuracy, with RuleFit nearly matching it and achieving the highest stability score. However, RuleFit’s performance began to show mild fluctuations in precision. Both random forest and SVM displayed wider spreads in violin Figure~\ref{fig:violin_std} and box plots and erratic line trends, indicating less reliable performance.

In the expanded dataset, SR4-Fit maintained high accuracy and rule stability, with consistent visual trends across all metrics. RuleFit showed increased sensitivity to feature complexity, reflected in the broader violin plot in Figure~\ref{fig:violin_exp} and fluctuating line plots, though it retained its interpretability advantage. Random forest remained unstable and uninterpretable, while SVM exhibited metric-dependent inconsistencies.

In the previous dataset, SR4-Fit continued to outperform the other models, further reinforcing its robustness and interpretability. RuleFit remained competitive, though with slightly increased variance, while random forest and SVM again demonstrated inconsistent performance and lacked interpretability. 

Statistical significance testing, including paired t-tests \cite{rouder2009bayesian} and p-value analysis \cite{biau2010p}, confirmed that the performance differences between SR4-Fit and RuleFit were generally statistically significant across datasets, as shown in Table~\ref{tab:ttests_split}, demonstrating that SR4-Fit’s improvements are not due to random variation but reflect a consistent and reliable advantage in both predictive accuracy and rule stability. The exception was with the previous dataset, which included information on previous election outcomes for each district, a very powerful predictor that leaves little room for more effective rules to be formed using demographic data.

These conclusions are further reinforced by the average interpretability scores reported in Table~\ref{tab:ips}, where SR4-Fit consistently achieved the highest IPS across all datasets, demonstrating its superior balance of predictive performance, model simplicity, and rule consistency. RuleFit showed strong interpretability and competitive IPS scores, particularly in the standard and expanded datasets, but was slightly outperformed by SR4-Fit across the board. In contrast, random forests recorded significantly lower IPS values due to their black-box nature and lack of rule-based transparency. The IPS violin plots, which can be seen in Figure~\ref{fig:ips_violin} visually reinforce these findings, with SR4-Fit consistently demonstrating higher IPS values with tighter distributions, indicating it produces more interpretable models with less variability. Overall, SR4-Fit emerged as the most dependable model across different models for demographic-based classification of U.S. House of Representatives elections, consistently delivering high accuracy, stable rules, and interpretable structures, even as feature complexity increases. 

\subsection{Interpretation SR4-Fit Rules for Election Forecasting}
\label{ss:SR4-Fit-interpretation}

The following analysis presents representative rules from each dataset and interprets the associated characteristics of regions these rules classify, illustrating how SR4-Fit balances prediction accuracy with interpretability.

In the minimum dataset, SR4-Fit builds a concise, interpretable model primarily driven by Democratic Percentage and Republican Percentage, the most influential features across all trials. The most consistently important rule was \textbf{DEM\_Percentage $>$ 48.7 and REP\_Percentage $\leq$ 45.5} (or variants with slightly different values), which indicates that districts in which registered Democrats notably outnumber registered Republicans are likely to elect Democrats, which is not surprising. When one party's registrations are much higher than the other's, a district is unlikely to be competitive.

Relatedly, \textbf{DEM\_Percentage $\leq$ 48.7 and REP\_Percentage $>$ 45.5}, typically signaled Republican elections. This may seem surprising at first because it could include districts in which registered Democrats slightly outnumber registered Republicans. However, it may be that the turnout rate of Republicans was higher than that of Democrats, that more Independents voted Republican than Democratic, or that more Democrats crossed over to vote for Republican candidates, at least in those districts in those years. 

This latter rule also received negative coefficients in several trials, indicating it predicted Democratic regions. In those exceptions, the regions had a median age of 39.24 years compared to the overall average of 37.49 years, a White non-Hispanic population share of 69.46\% compared to 63.64\%, and a bachelor's degree attainment rate of 19.06\% compared to 17.91\%. In other words, even though regions without a notable Democratic majority often elect Republicans to the U.S. House, these districts where people were, on average, older, whiter, and had more education often elected Democrats. These demographic traits are widely understood to influence voting patterns.

In the standard dataset, SR4-Fit builds an interpretable model shaped by age distribution, party support, and socioeconomic indicators. Age between 45 and 54 emerged as the most influential feature, consistently receiving high positive coefficients, indicating Republican-leaning tendencies in middle-aged dominant regions. Similarly, age between 65 and 74 and age between 18 and 24 also showed strong positive associations with Republican classification, highlighting the predictive value of both older and younger age groups. 

A comparatively complex rule, \textbf{DEM\_Percentage $>$ 48.33 and REP\_Percentage $\leq$ 49.99 and DEM\_Percentage $\leq$ 48.70 and Median\_Income $\leq$ 23,717.5} had minimal positive influence, suggesting that low income may subtly shift predictions toward Republican classification in closely contested regions.\footnote{Note that we report rules as they are created by SR4-Fit. For human interpretation, it would make sense to slightly rewrite them. For example, this rule could be rewritten as \textbf{48.33 $<$ DEM\_Percentage $\leq$ 48.70 and REP\_Percentage $\leq$ 49.99 and Median\_Income $\leq$ 23,717.5}.} 

Additionally, a narrow-range rule \textbf{DEM\_Percentage $\leq$ 48.70 and REP\_Percentage $>$ 43.35 and DEM\_Percentage $\leq$ 47.96 and REP\_Percentage $\leq$ 45.45 and Bachelor\_s $>$ 21.7} received a small positive coefficient, suggesting that a relatively high educational attainment rate may nudge moderately split districts toward Republican classification.

In the expanded dataset, SR4-Fit captures nuanced voting behavior through interpretable rules primarily driven by Democratic and Republican support percentages. One such rule, \textbf{DEM\_Percentage $\leq$ 48.33 and REP\_Percentage $>$ 43.40 and REP\_Percentage $\leq$ 46.97 and DEM\_Percentage $>$ 46.31}, received a negative coefficient, signaling a tendency toward Democratic classification in moderately split districts. The districts satisfying this rule and resulting in Democratic outcomes had a median age of 38.97 years compared to the overall average of 37.49 years, a higher Asian population share of 8.48\% versus 4.86\%, and a high multiracial share of 3.95\% compared to 2.68\%. These regions also had a higher White non-Hispanic population of 67.70\% compared to 63.64\%, a higher bachelor's degree attainment rate of 19.38\% versus 17.91\%, a higher median income of \$30,764.50 compared to \$26,873.71, and lower poverty levels of 11.22\% compared to 14.63\%.

In the previous dataset, SR4-Fit identifies the rule \textbf{DEM\_Percentage $>$ 48.7147 and REP\_Percentage $>$ 49.9979} as a strong positive coefficient signaling for Republican classification, highlighting districts where both parties have high support but Republicans narrowly dominate. The districts satisfying this rule and resulting in Republican outcomes had a slightly older population, with a median age of 37.70 years compared to the overall average of 37.49 years. These regions were notably less diverse, with a higher White non-Hispanic population share of 71.58\% versus 63.64\%, and lower representation of Black (8.25\% vs. 12.58\%), Asian (3.40\% vs. 4.86\%), Hispanic (14.41\% vs. 16.27\%), and multiracial (2.27\% vs. 2.68\%) groups. Educational attainment was also higher in these districts, with 19.37\% holding a bachelor's degree and 11.68\% holding a graduate or professional degree, compared to 17.91\% and 10.65\%, respectively, across the dataset. 

Based on SR4-Fit’s interpretable rules across all datasets, Democratic classifications are generally associated with higher racial diversity—particularly greater Asian, Hispanic, and multiracial populations, along with moderate to high Democratic support, elevated educational attainment, younger populations, and, in some cases, higher median income and lower poverty levels. In contrast, Republican classifications are more common in districts with higher White non-Hispanic population shares, older age groups (especially 45–54 and 65–74), and occasionally higher educational attainment, particularly in less diverse regions. Even in tightly split districts, small advantages in Republican support combined with lower diversity often shift the outcome. These findings suggest that SR4-Fit captures not only the support margins but also the demographic and socioeconomic contexts that shape political alignment.

\subsection{Result Analysis on Standard Public Datasets}
\label{ss:result-public}

Across all the standard datasets used for experimentation, SR4-Fit consistently demonstrates competitive performance compared to RuleFit and random forest, supported by accuracy, stability, and interpretability evaluations. As reflected in the accuracy, precision, recall, and F1-score plots (see Supplementary Section B), SR4-Fit frequently matches or exceeds RuleFit in predictive performance while maintaining far greater consistency across trials. For instance, in the Breast Cancer dataset, SR4-Fit achieves a high accuracy of 0.9650, closely followed by RuleFit, with the lowest variance visible in both the line plots and the narrow violin and box plot distributions, indicating strong generalization and stability. The Vehicle dataset shows a similar trend, where SR4-Fit maintains 0.9565 accuracy, competitive with the highest-performing model (SVM at 0.9741), while demonstrating far more stable rule behavior across trials.

These performance patterns are mirrored in the Interpretability Score (IPS) plots, where SR4-Fit again outperforms other rule-based models in Breast Cancer (0.5875), Ecoli (0.5924), and PageBlocks (0.4570), reflecting not just high accuracy but also superior rule stability and simplicity. RuleFit follows closely in several datasets, especially Vehicle, where it achieves the highest IPS (0.5151), but exhibits wider violin spreads and more variability in line plots, indicating slightly less robustness. Random forest, although occasionally showing high raw accuracy (e.g., 0.9088 in Ecoli), consistently lags in interpretability, with the lowest IPS values in most datasets and more erratic behavior in accuracy and metric plots. Notably, in the Pima Indians dataset, random forest attains the highest IPS (0.4665) among the rule-based models, suggesting it may benefit from simpler feature spaces, while both RuleFit and SR4-Fit show significant drops (IPS of 0.2050 and 0.1833, respectively), potentially due to overfitting or instability in limited-dimensional settings.

In Yeast, a complex and imbalanced dataset, SR4-Fit still performs competitively in both accuracy and IPS (0.4835), slightly trailing random forest but maintaining a tighter and more reliable spread in its violin plots. Overall, the combination of high accuracy, minimal variance across trials, and consistently superior interpretability scores makes SR4-Fit the most dependable and generalizable rule-based model across datasets (see supplementary section C). It balances both performance and transparency, with RuleFit serving as a strong but slightly less stable alternative, while random forest, despite occasional IPS spikes, remains less interpretable and more variable across metrics.

\section{Conclusions and Future Work}
\label{sec:Conclusion}
The SR4-Fit algorithm addresses the balance between interpretability and predictive performance by combining rule-based modeling with sparse optimization. It performs comparably to traditional interpretable models like RuleFit and black-box models such as random forest while producing concise and stable rules that can be understood by users. In addition to its application to U.S. House election data, SR4-Fit shows reliable behavior across various standard public datasets, including Breast Cancer, Ecoli, Page Blocks, Vehicle, Pima Indians Diabetes, and Yeast, maintaining consistent accuracy and interpretability.

The SR4-Fit model also highlights demographic trends, where Democratic classifications are associated with greater racial diversity, younger populations, higher educational attainment, and, in some cases, higher income and lower poverty, while Republican classifications are observed in districts with higher proportions of White non-Hispanic populations, older age groups, and lower diversity. These findings contribute to both prediction and understanding of demographic influences on electoral outcomes.

Future work can explore the use of SR4-Fit in domains that involve complex relationships of the structured data and require interpretable modeling for decision-making and planning. Extending SR4-Fit to handle regression tasks would enable the prediction of continuous outcomes while maintaining transparency.

\balance

\bibliographystyle{IEEEtran}
\bibliography{bibliography.bib}
\end{document}